\documentclass[10pt, a4paper]{article}

\usepackage[]{lrec-coling2024} 
\usepackage[normalem]{ulem}
\useunder{\uline}{\ul}{}
\usepackage[inline]{enumitem}
\usepackage{enumitem}

\title{Domain Generalization via Causal Adjustment for \\Cross-Domain Sentiment Analysis}

\name{\normalsize Siyin Wang$^1$, Jie Zhou$^{2,*}$\thanks{$^*$ Corresponding authors, jzhou@cs.ecnu.edu.cn, xjhuang@fudan.edu.cn.}, Qin Chen$^2$, Qi Zhang$^{1}$, Tao Gui$^3$, Xuanjing Huang$^{1,*}$} 

\address{\normalsize $^1$ School of Computer Science, Fudan University, Shanghai, China \\
         $^2$ School of Computer Science and Technology, East China Normal University, Shanghai, China \\
         $^3$ Institute of Modern Languages and Linguistics, Fudan University, Shanghai, China \\
}

\abstract{
Domain adaption has been widely adapted for cross-domain sentiment analysis to transfer knowledge from the source domain to the target domain. Whereas, most methods are proposed under the assumption that the target (test) domain is known, making them fail to generalize well on unknown test data that is not always available in practice. 
In this paper, we focus on the problem of domain generalization for cross-domain sentiment analysis. 
Specifically, we propose a backdoor adjustment-based causal model to disentangle the domain-specific and domain-invariant representations that play essential roles in tackling domain shift. 
First, we rethink the cross-domain sentiment analysis task in a causal view to model the causal-and-effect relationships among different variables.
Then, to learn an invariant feature representation, we remove the effect of domain confounders (e.g., domain knowledge) using the backdoor adjustment.
A series of experiments over many homologous and diverse datasets show the great performance and robustness of our model by comparing it with the state-of-the-art domain generalization baselines. 
The codes of our model and baselines are available at \href{https://github.com/sinwang20/DeepDG4nlp}{https://github.com/sinwang20/DeepDG4nlp}.
 \\ \newline  }

\begin{document}

\maketitleabstract

\section{Introduction}
\label{sec:intro}
\begin{figure}[!ht]
    \centering
    \includegraphics[scale=0.28]{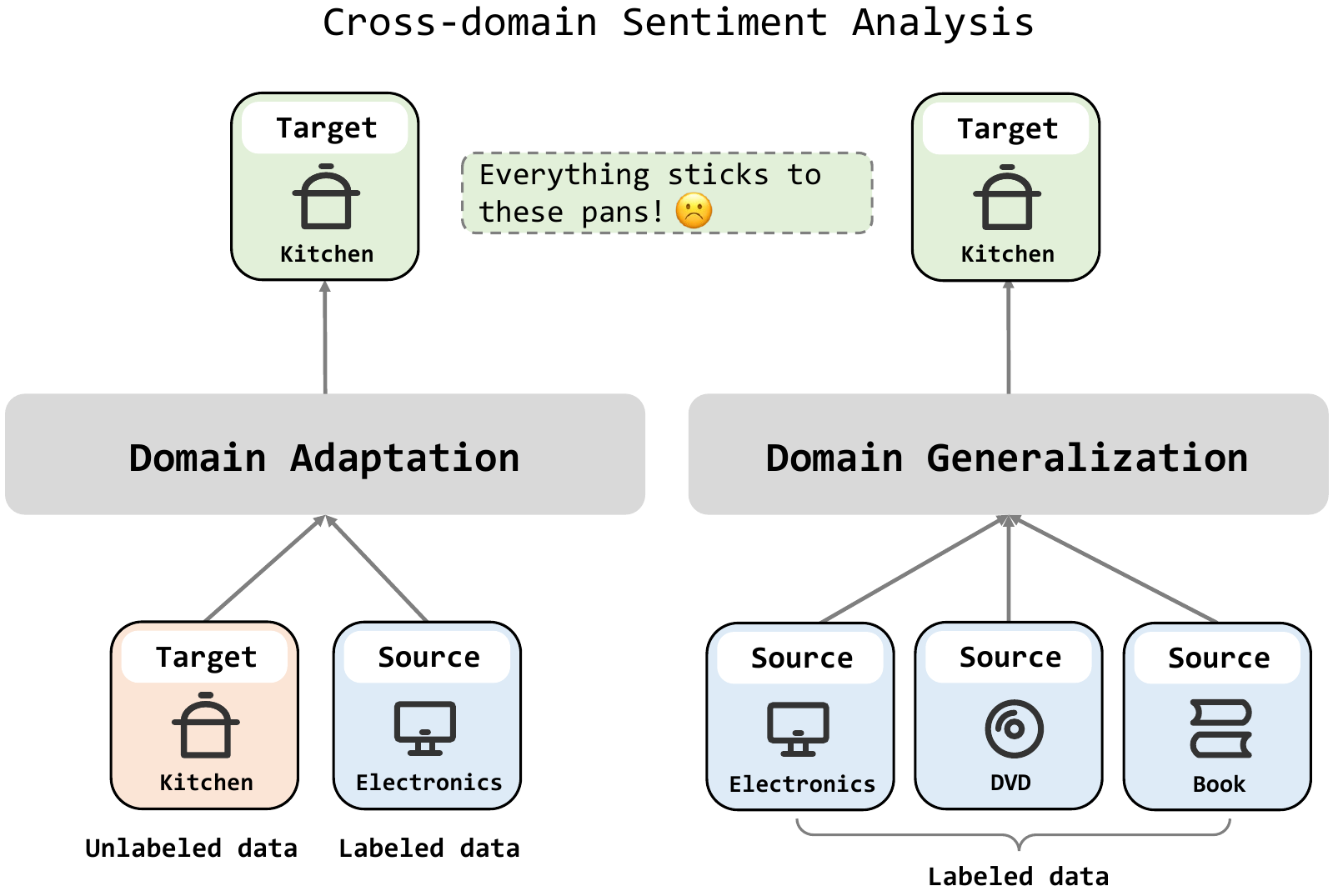}
    \caption{Difference between domain adaptation and domain generalization.}
    \label{fig:intro}
\end{figure}

In the field of cross-domain sentiment analysis \cite{zhou2020sentix,du2020adversarial}, domain adaptation (DA) has been extensively studied to transfer the sentiment knowledge from a source (label-rich) domain to a target domain. However, most existing methods in this area assume that the target domain is known during the training phase, which limits their generalization performance when applied to unknown test domain, a scenario commonly encountered in practical applications. 
In reality it is often, such as in Amazon's products, collecting unlabeled data and fine-tuning (like domain adaptation) is expensive and extravagant, prohibitively impossible. 
To address this problem, we focus on Domain Generalization (DG) in the field of cross-domain sentiment analysis, which sets a more strict situation, the test domain is unseen (Figure \ref{fig:intro}).

Recently, domain generalization \cite{wang2022generalizing} has attracted increasing interest in the field of computer vision, which aims to learn a model that can generalize to an unseen target domain from some different but related source domains.

The existing methods mainly focus on learning general invariant representations from multiple domains by data manipulation \cite{adila2022understanding,volpi2018generalizing}, adversarial training \cite{ganin2016domain,arjovsky2019invariant}, and meta-learning \cite{chen2022discriminative}. 

However, there are two significant challenges with current domain generalization methods for cross-domain sentiment analysis.
One major challenge (\textbf{C1}) is the existence of spurious correlations in domain invariance. 
It is hard to guarantee that the learned representation is the true cause of the sentiment polarity. 
For example, the term "hot" indicates popularity in book domain (e.g., ``hot-selling books") and delicacy in kitchen domain (e.g., ``hot pizza"), leading to the invariant behavior across domains. This spurious correlation fails to hold in other domains, such as ``hot CPU," where the sentiment may differ. Such false invariances can degrade the performance of sentiment analysis models in the presence of domain shift.

Furthermore, another challenge (\textbf{C2}) is that many existing models focus solely on capturing domain-invariant information, such as general sentiment words (e.g., good, bad), while disregarding crucial domain-specific knowledge. This approach results in the loss of valuable domain-specific features that are essential for accurate sentiment analysis. For the previous example, if a model learns that ``hot" represents a successful book sale in the book domain and ``hot" represents terrible in electronics domain, it can understand that ``hot DVD" is positive, while "hot DVD device" is negative, leveraging the domain-specific information.

To address these challenges, we first rethink the task of cross-domain sentiment analysis from a causal perspective, aiming to model the causal relationships among different variables. 
Then, we propose a causal adjustment-based framework for domain generalization, which learns the generalized representation by disentangling the domain-invariant and domain-specific information.
Our model shows great performance over both homologous and diverse datasets. 
We also explore the robustness of our model on 13 unseen homologous datasets and do ablation studies to verify the effectiveness of components consisting of our model. 
Representation visualization shows that our model learns a better domain-invariant representation than the baseline.

Our key contributions are summarized as follows.
\begin{itemize}[leftmargin=*, align=left]
    \item We focus on domain generalization for cross-domain sentiment analysis with the assumption that test domains are unseen. Moreover, we rethink this task in a causal view to analyze causal-and-effect relationships among various variables.
    
    \item We propose a causal adjustment method to disentangle the domain-invariant and domain-specific representation.
    
    \item Extensive experimental results on more than 20 homologous and diverse datasets indicate that our model can remove the influence of confounders to learn a generalized representation.  
\end{itemize}

\section{Preliminaries}
\label{sec:prelim}
\subsection{Formulation}
\label{sec:formulation}

We define cross-domain sentiment analysis as follows. In the training phase, we have datasets $ {\mathcal{D}^{train}} = \{(x_i^d, y_i^d)\}_{i=1}^{N_d}, d \in \{1,2,..\xi\}$, where $x_i^d$ denotes the $i$th input text(training sample) from the $d$th source domain, $y_i^d$ is the corresponding sentiment label, and $N^d$ is the number of training samples in domain $d$.
The goal of domain generalization is that given a sample $x^T$ from an unseen domain, we aim to predict its output $\hat{y}^T$ through generalizable representation $\Phi(x)$.

Unlike traditional domain adaptation methods that align representations between source and target domains or other methods that focus on finding invariant representations $\Phi(x_{inv})$, we propose a novel approach that considers both domain-invariant and domain-specific representations based on causal mechanisms and achieve a better generalizable representation $\Phi(x)$. 

\subsection{Structural Causal Model}
\label{sec:scm}

{Structural Causal Models (SCMs)} \cite{pearl2000models} are widely used to represent causal relationships, such as the causal relationship between the text $X$ and sentiment $Y$. They are depicted as directed acyclic graphs (DAGs) $G = \left\{V, E\right\}$, where $V$ represents the set of variables (e.g., text $X$, sentiment $Y$, domain $D$) and $E$ represents the direct causal connections.

In our work, we utilize SCMs to model the relationships among variables in cross-domain sentiment analysis by specifying how the value of a variable is determined given its parents. These relationships are known as Causal Mechanisms \cite{peters2017elements}.
Specifically, in our model, the sentiment $Y$ is influenced by its parental variables, which consist of the domain-invariant factor $X_{pa(Y)}^{inv}$ and the domain-specific factor $X_{pa(Y)}^{spc}$. We represent this relationship as follows:

\textbf{Definition 1} \textit{(Causal Mechanisms)}
$$ Y \leftarrow f_{Y} (X_{pa(Y)}^{inv}, X_{pa(Y)}^{spc} , \epsilon_Y ),  X_{pa(Y)}^{inv} \perp \!\!\! \perp \epsilon_Y$$
 
Here, $pa(Y)$ refers to the set of parental variables for sentiment $Y$. The parental set includes both the domain-invariant factor $X_{pa(Y)}^{inv}$ and domain-specific factor $X_{pa(Y)}^{spc}$ of sentiment $Y$. $\epsilon_Y$ represents the errors due to omitted factors.

In simpler terms, our model captures the causal relationships between the variables. The sentiment $Y$ is influenced by both the domain-invariant aspects of the text $X$ and the domain-specific characteristics. By considering these causal relationships, we can better understand and analyze the sentiment in cross-domain scenarios.

\subsection{Backdoor Adjustment}
\label{sec:backdoor adjustment}

We first consider a simplified setting with only text ($X$), sentiment labels ($Y$) and confounders ($D$).
In sentiment analysis, where the goal is to predict sentiment labels (Y) based on text inputs (X), it is important to consider potential confounders that may introduce biases and shortcuts in the causal relationships. One such confounder is the presence of domains (D), which can influence both the text inputs and the sentiment labels. 

To tackle the potential confounders existing in the causal inference, one of the regular methods is backdoor adjustment \cite{pearl2000models} (\textbf{Definition 2}). This approach identifies the pure causal effect $P(Y \mid do(X))$ from the total effect $P(Y \mid X)$ by eliminating the supurios correlation of potential backdoor paths, i.e. $X \leftarrow D \rightarrow Y$.

\textbf{Definition 2} \textit{(Backdoor Adjustment Formula)}
$$ P (Y \mid do(X)) = \sum_{D} P (Y \mid X, D)P (D) $$

Through this adjustment, we can get the pure relationship between $X$ and $Y$, $X \rightarrow Y$ without the backdoor path, $X \leftarrow D \rightarrow Y$. 

The backdoor adjustment formulation cannot be used arbitrarily, the adjustment variable $D$ should satisfy the backdoor criterion relative to $X$ and $Y$ if:

\begin{itemize}
\item[$\bullet$] No node in $D$ is a descendant of $Y$.
\item[$\bullet$] Every path between $X$ and $Y$ that contains an arrow pointing to $X$ is blocked by $D$.
\end{itemize}

In our SCM, we consider the adjustment between $M_{inv}$ and $Y$, and the adjustment varivable is the domain $D$.
Fortunately, the domain $D$ satisfies the backdoor criterion: (1) sentiment label $D$  obviously cannot be the factor of domain and therefore is not the descendant of $Y$. (2) "path between $M_{inv}$ and $Y$ that contains an arrow pointing to $M_{inv}$" is the backdoor path between $M_{inv}$ and $Y$. The two backdoor path in our SCM, i.e. $M_{inv} \leftarrow D \rightarrow Y$ and $M_{inv} \leftarrow D \leftarrow M_{spc} \rightarrow Y$ both can be blocked by $D$. It means that when condition on $D$ as the backdoor adjustment formula, $D$ elimate the correlation between $M_{inv}$ and $Y$. 

So the causal effect between $M_{inv}$ and $Y$ is identifiable and can be formulated as,
$$ P (Y \mid do(M_{inv})) = \sum_{D} P (Y \mid M_{inv}, D)P (D) $$
From the adjustment above, The pure causal effect $P (Y \mid do(M_{inv}))$ is extracted from the total effect $P (Y \mid M_{inv})$ by removing the spurious correlation caused by backdoor path.

\section{Our Approach}
\label{sect:isaiv}
We propose a backdoor adjustment-based causal model for cross-domain sentiment analysis. 
We first rethink this task in a causal view (Section \ref{sect:Causal View of Cross-Domain Sentiment Analysis}).
Then, we introduce the overview of our model (Section \ref{sect:overview}) and integrate backdoor adjustment to learn a better invariant representation (Section \ref{sect:Invariant Representation Learning via Backdoor Adjustment}).

\subsection{Causal View of Cross-Domain Sentiment Analysis}
\label{sect:Causal View of Cross-Domain Sentiment Analysis}

\begin{figure}
    \centering
    \includegraphics[scale=0.4]{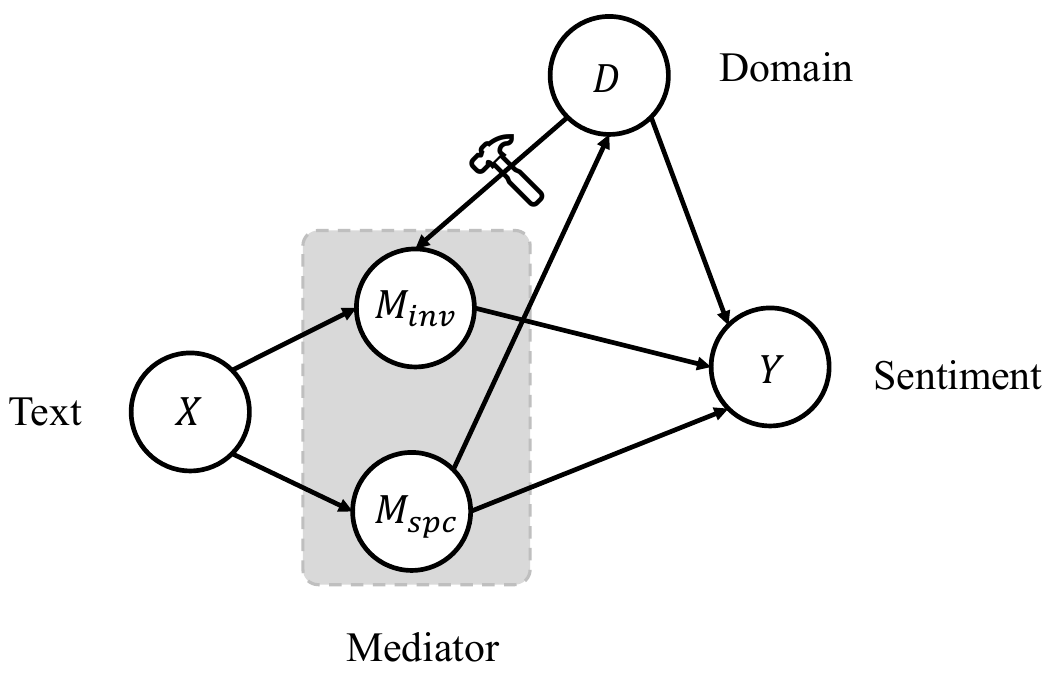}
    \caption{Structural Causal Model of Cross-domain Sentiment analysis}
    \label{fig:causal structure}
\end{figure} 

Despite achieving great performance under the i.i.d condition, the model will fail when encountering the problem of domain shift. 
That's because the training goal is to minimize $\mathbb{E}_{(x,y)\sim P^\mathrm{train}(X,Y)} l(f(x), y)$, where $f$ represents the model, and the objective is to minimize the loss $l$ between the model's predictions and the true labels on the training data. But the empirical distribution of training data $P^\mathrm{train}(X, Y)$ is not identical to the test data distribution $P^\mathrm{test}(X, Y)$.
As the $P(X,Y) = P(Y|X)P(X)$, $P(X)$ is the marginal distribution of $X$. The period work under the assumption that $P(Y|X)$ remains stable, expected to align the $P^\mathrm{train}(X)$ to $P^\mathrm{test}(X)$, like many domain adaption methods to align the representation using unlabeled test domain data to achieve the better cross-domain performance \cite{ganin2016domain, du2020adversarial}.
Different from DA, the test data is unseen in the DG setting, which makes the prior distribution of target domain $P^\mathrm{test}(X)$ cannot be accessed.

\begin{figure*}[t!]
\centering
\includegraphics[scale=0.4]{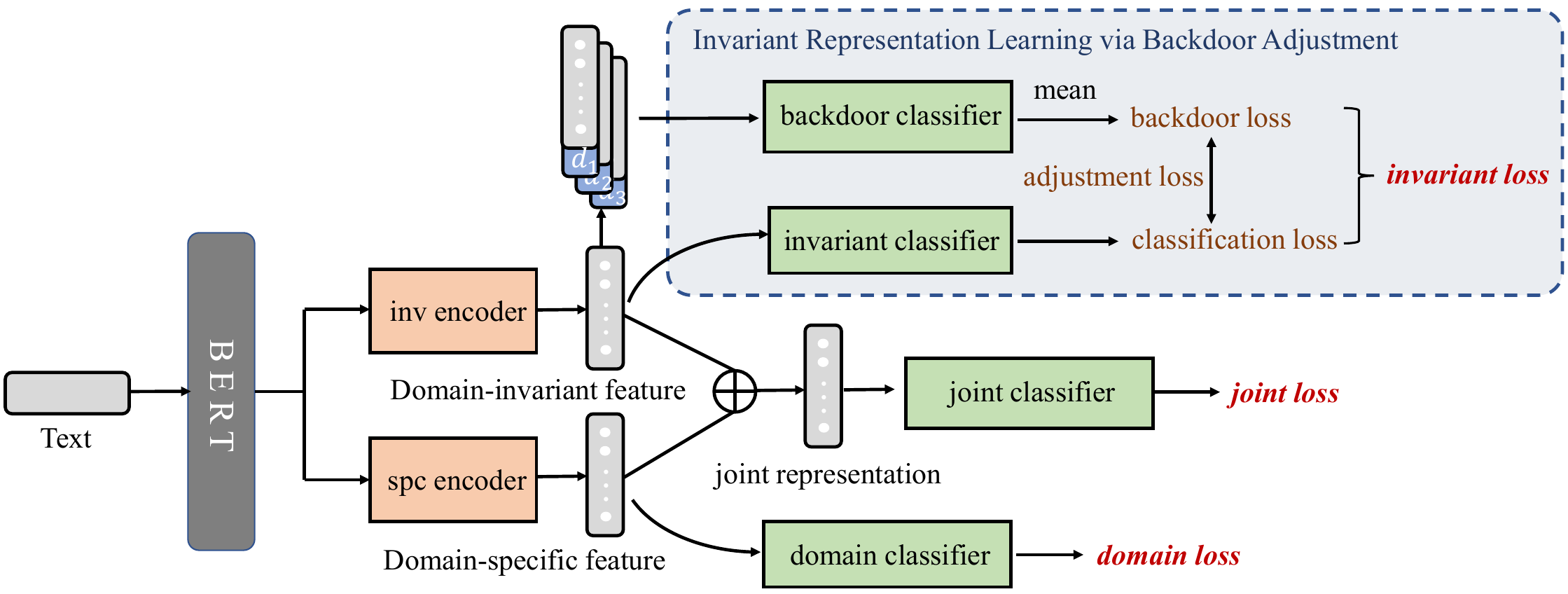}
\caption{The framework of our model.}
\label{fig:framework}
\end{figure*}

The key challenge for domain generalization is to learn a generalizable representation $\Phi(X)$ without the test domain distribution, which performs well over all domains.
Fortunately, the causal mechanisms, $P(Y|\Phi(X))$ has the ability to generalize to the unseen target domain \cite{Scholkopf2021TowardCR}.

We then construct the Structural Causal Models of cross-domain sentiment analysis in Figure \ref{fig:causal structure} to illustrate the causal relationship with the variables we used. Along with the causal mechanism in section \ref{sec:scm}, we disentangle the text input into domain-invariant and -specific features to model the causal mechanism between text and sentiment.
These two features both will cause the sentiment, denoted as the path $M_\mathrm{inv} \rightarrow Y$ and $M_\mathrm{spc} \rightarrow Y$. Note that we also consider the relationship $M_\mathrm{spc} \rightarrow Y$ which is different from the generalization in image classification, because the domain-specific information like the word ``hot" in different domains will also affect the sentiment.
For domain generalization, another critical variable is the Domain variable $D$, which represents the domain of the text. Obviously, the domain-specific feature is the cause of the domain. 
Additionally, the domain variable serves as a confounder that affects the prediction of sentiment $Y$ as well as text.
An inexhaustive disentanglement of domain-invariant and -specific may cause the $M_\mathrm{inv}$ to suffer from the effect of the domain variable.

\subsection{Overview of Proposed Model}
\label{sect:overview}
We introduce the structure of our causal model for cross-domain sentiment analysis (Figure \ref{fig:framework}). 
We use a BERT $\mathcal{M}$ model to obtain the representation of text $x$ using the ``[CLS]" representation.
\begin{equation}
    h = \mathcal{M}(x)
\end{equation}
where $h$ is the text representation. 
Then, to extract the domain-invariant and domain-specific features, we set two independent encoders $\phi$ (invariant encoder and specific encoder), a three-layer multi-layer perception (MLP) with ReLU activation, $m_\mathrm{inv} = \phi_\mathrm{inv}(h), m_\mathrm{spc} = \phi_\mathrm{spc}(h)$.

To help learn the domain-specific information in multiple domains, we constrain its learning with the domain classification loss.
\begin{equation}
    \mathcal{L}_\mathrm{specific} = 
    \mathrm{CE}( f_\mathrm{specific}(m_\mathrm{spc}), d )
\end{equation}
where $d$ denotes the domain of $m_\mathrm{spc}$.

As both the domain-invariant and -specific features are the causal factors of the output (polarity), the joint representation ($m_{joint}$) is formed through element-wise addition of the domain-invariant feature ($m_{inv}$) and the domain-specific feature ($m_{spc}$). To ensure accurate polarity prediction, we employ a cross-entropy loss function, which facilitates the alignment between the joint representation and the target sentiment label ($y$) . 
\begin{equation}
    \mathcal{L}_\mathrm{joint} = 
    \mathrm{CE}( f_\mathrm{joint}(m_\mathrm{inv} + m_\mathrm{spc}), y)
\end{equation}
where ``+" means add by elements.

For a domain-invariant representation, we design the invariant loss $\mathcal{L}_\mathrm{invariant}$ from the standpoint of causality and will elaborate on it in Section \ref{sect:Invariant Representation Learning via Backdoor Adjustment}.

Overall, our loss function is as follows,
\begin{equation}
     \mathcal{L}_\mathrm{all} = 
 \mathcal{L}_\mathrm{joint} + \mathcal{L}_\mathrm{invariant} + \mathcal{L}_\mathrm{specific} 
\end{equation}


\subsection{Invariant Representation Learning via Backdoor Adjustment}
\label{sect:Invariant Representation Learning via Backdoor Adjustment}

\paragraph{Theoretical Analysis} From a causal perspective, if the invariant representation wrongly contains some specific information as the inadequate disentanglement, a path $D \rightarrow M_\mathrm{inv}$ emerges that should not exist. Which means that the domain variable affects the wrong invariant representation and leads to the spurious correlation between $M_\mathrm{inv}$ and $Y$, denoted as $M_\mathrm{inv} \leftarrow D \rightarrow Y$.

For example, model may wrongly think ``hot'' is an invariant sentiment word, but the relationship between ``hot'' and sentiment is caused by the spurious correlation by domain. We are more likely to describe the heating capacity of kitchen application with ``hot" instead of ``warm" (a common word in the clothes domain), like \textit{``hot enough to boiling water"}, ($D \rightarrow M_\mathrm{inv}$).
Furthermore, in the electronics domain, "hot" is more likely to be associated with negative polarity, such as "a very hot CPU," indicating a relationship between the invariant representation and sentiment, i.e., $M_\mathrm{inv} \leftarrow D \rightarrow Y$.




To address the issue of spurious correlations, we can adjust them using the Backdoor Adjustment Formula (Definition 2) and identify the backdoor path involving $D$ conventionally. 
Given the test domain is unknown, we propose incorporating a constraint condition during the training phase to achieve a purer invariant representation instead of adjusting the correlations during inference.

In our specific task, we consider that if the learned invariant features are truly domain-invariant, then the domain should have no effect on these features (i.e., $M_\mathrm{inv}$ is independent of $D$). 
Therefore, the prediction probabilities, with and without backdoor adjustment, should be identical, referred to as "Backdoor Condition" we proposed.



\textit{Proposition of Backdoor Condition} 
If a learned representation $M$ is an invariant representation, then the backdoor adjustment is invalid for it, i.e. $P(Y \mid do(M=m)) = P(Y \mid M=m)$.

\textit{Proof of Backdoor Condition} 
The invariant representation obviously should be independent of the domain variable, denoted as $M \perp \!\!\! \perp D$. According to the backdoor adjustment formula, $P(Y \mid do(M=m) = \sum_{D}P (Y \mid X, D)P (D) $. And the conditional distribution of Y given X can be written as, $P(Y \mid M=m) = \sum_{D} P (Y \mid X, D)P (D \mid M)$. As the $M \perp \!\!\! \perp D$, so the $P(Y \mid do(M=m) = P(Y \mid M=m)$.

By incorporating the Backdoor Condition into the training phase, we aim to facilitate effective disentanglement between domain-invariant and -specific features. 
This approach disentangles the backdoor and causal paths during training rather than during inference, as the target domain is unknown.
\paragraph{Loss Design}
In practice, we design an adjustment loss to achieve the Backdoor Condition to approximate invariant representation. First, we design the after-adjustment distribution of $M_\mathrm{inv}$ according to the backdoor adjustment (section \ref{sec:backdoor adjustment}), and set the corresponding loss to help the construction.
\begin{equation}
\small
\nonumber
    \mathcal{L}_\mathrm{backdoor} = 
    \mathrm{CE}( \sum_{d \in D}{f_\mathrm{backdoor}(m_\mathrm{inv} \oplus e^d) \cdot P(d)}, y )
\end{equation}
where $e^d$ is set as the learnable embedding of domain $d$ and $\oplus$ means concatenation. We simplify $P(d)$ to the proportion of the domain $d$ in the input data, i.e. $P(d) = \frac{1}{\mid D \mid}$.

The $P (Y \mid M_{inv})$ is modeled by the classification loss,
\begin{equation}
    \mathcal{L}_\mathrm{classification} = 
    \mathrm{CE}( f_\mathrm{classification}(m_\mathrm{inv}), y )
\end{equation}

Adjustment loss aligns the $P(Y \mid M_\mathrm{inv})$ and $P(Y \mid do(M_\mathrm{inv}))$ to achieve the invariant representation under the backdoor condition.
\begin{equation}
\nonumber
    \begin{aligned}
    \mathcal{L}_\mathrm{adjustment} &= {\mid \mathcal{L}_\mathrm{classification} - \mathcal{L}_\mathrm{backdoor} \mid}^2 \\
    &= [\sum_{i=1}^C{y_i \log(y^\mathrm{inv}_i)}-\sum_{i=1}^C{y_i \log(y^\mathrm{back}_i)}]^2 \\
    &= [\sum_{i=1}^C{y_i \log(\frac{y^\mathrm{inv}_i}{y^\mathrm{back}_i})}]
    \end{aligned}
\end{equation}

The above derivation shows that such a loss setting makes the $P(Y \mid M_\mathrm{inv})$ equals to $P(Y \mid do(M_\mathrm{inv}))$.

Finally, we set the $\alpha$ and $\beta$ to adjust the weight of the backdoor classifier and adjustment. The complete invariant loss is as follows,
\begin{equation}
\begin{split}
     \mathcal{L}_\mathrm{invariant} = 
\mathcal{L}_\mathrm{classification}
 + \alpha \cdot \mathcal{L}_\mathrm{backdoor} \\ + \beta \cdot {\mid \mathcal{L}_\mathrm{invariant} - \mathcal{L}_\mathrm{backdoor} \mid}^2
\end{split}
\end{equation}


\section{Experimental Setups}
\label{sec:experimental setups}

\subsection{Datasets and Metrics}
\label{sect:datasets and metrics}
\paragraph{Homologous Datasets} We use the multi-domain Amazon reviews dataset \citeplanguageresource{blitzer-etal-2007-biographies}, a widely-used standard benchmark datasets for domain adaptation. It contains reviews on four domains: Books (B), DVDs (D), Electronics (E), and
Kitchen appliances (K). For domain generalization, We follow the experiment settings proposed by \citeplanguageresource{ziser-reichart-2017-neural}. Each domain also has 2,000 labeled examples (1,000 positive and 1,000 negative).
To further evaluate our model's performance and robustness, we adopt dozens of unseen datasets of Amazon reviews dataset \citeplanguageresource{blitzer-etal-2007-biographies}, with 13 types of products. 

\paragraph{Diverse Datasets}
To consider a more challenging setup we experiment with a large gap domain generalization. We randomly sample the 2,000 labeled examples (1,000 positive and 1,000 negative) from four sentiment analysis datasets, including products domain from Amazon reviews \citeplanguageresource{blitzer-etal-2007-biographies}, restaurant domain from Yelp reviews \citeplanguageresource{NIPS2015_250cf8b5}, airline domain from airline reviews \footnote{\href{https://github.com/quankiquanki/skytrax-reviews-dataset}{https://github.com/quankiquanki/skytrax-reviews-dataset}}, movie domain from IMDb reviews \citeplanguageresource{maas-etal-2011-learning}.
In contrast to Homologous Datasets, Diverse Datasets are extracted from different writing platforms and therefore are more diverse in terms of the writing type, navigator, etc.

\subsection{Baselines}
\label{sect:baselines}
As most of the past research in cross-domain sentiment analysis concentrates on domain adaptation, which requires the target domain data, we mainly compare our proposed model with several popular and strong domain generalization methods.
\textbf{(1) MoE, MoEA} \cite{guo-etal-2018-multi} model the domain relationship with a mixture-of-experts (MoE) approach in non-adversarial and adversarial settings.
\textbf{(2) BERT-base (ERM)} 
    is a basic BERT model with a binary classification layer at the output and minimizes empirical risk.
\textbf{(3) DANN} \cite{ganin2016domain} utilizes an adversarial approach to learn features to be domain indiscriminate.
\textbf{(4) Mixup} \cite{DBLP:conf/iclr/ZhangCDL18, DBLP:conf/coling/SunXYLYH20} adopts pairs of examples from random domains along with interpolated labels to perform ERM.
\textbf{(5) GroupDRO} \cite{sagawa2019distributionally} explicitly minimizes the loss in the worst training environment to tackle the problem that the distribution minority lacks sufficient training.
\textbf{(6) IRM} \cite{arjovsky2019invariant} seeks data representations where the optimal classifier on top of those representations matches across randomly partitioned environments.
\textbf{(7) VREx} \cite{DBLP:conf/icml/KruegerCJ0BZPC21} reduces the variance of risks in test environments by minimizing the risk variances in training environments.
\textbf{(8) EQRM} \cite{Eastwood2022ProbableDG} leverages invariant risk like VREx, but learns predictors that perform well with high probability rather than on-average or in the worst case.

\subsection{Implementation Details} 
\label{sect:implementation details}
In accordance with the commonly used leave-one-domain-out protocol \cite{DaLi2017DeeperBA}, one domain will be set aside for testing and the remaining domains will be used for training. The data in each training domain is randomly divided into a training set (80 \%) and a validation set (20 \%).
During training, the learning rate is set as 1e-5 and the batch size is set as 16. Adam optimizer \cite{kingma2014adam} is used to update all the parameters. For our $\mathcal{L}_\mathrm{invariant}$
, $\alpha $ and $\beta$ are searched in $[0.1, 100]$.
For the representation visualization, both settings are identical except for the input variations, with n\_components=2 and perplexity=100.

\section{Experimental Analysis}
In this section, we first evaluate the performance (Section \ref{sect:main results}) and robustness (Section \ref{sect:robustness}) of our model by comparing it with baselines. 
Then, we conduct experiments on diverse domains to further verify the model's effectiveness (Section \ref{sect:Diversity Domain}). 
Finally, we report more analysis on ablation studies, representation visualization and comparison with Large Language Models (Section \ref{sect:further analysis}).

\subsection{Main Results} 
\label{sect:main results}
\begin{table}[t!]
\centering
\small
\setlength{\tabcolsep}{0.8mm}{\begin{tabular}{lccccc}
\hline
          & DEK-B          & EKB-D & KBD-E          & BDE-K & Avg            \\ \hline
MoE       & 87.55          & 87.85 & 89.20          & 90.45 & 88.76          \\
MoE-A     & 87.85          & 87.65 & 89.50          & 90.45 & 88.86          \\
Bert-base & 88.10          & 89.65 & 90.00          & 90.35 & 89.53          \\
DANN      & 88.80          & 89.75 & 89.80          & 89.95 & 89.58          \\
Mixup     & 87.40          & 89.20 & 89.00          & 89.95 & 88.89          \\
GroupDRO  & 89.65          & 89.65 & 89.20          & 89.75 & 89.56          \\
IRM       & 88.55 & 90.05 & 88.80 & 91.00 & 89.60       \\ 
VREx      & 88.40          & 89.80 & 90.30          & 90.35 & 89.71          \\ 
EQRM      &  88.55	   & 89.80       &  90.40       &  90.85              & 89.90      \\ 
\hline
Ours      & \textbf{90.20} & \textbf{90.15} & \textbf{90.95} & \textbf{91.95} & \textbf{90.81} \\
\hline
\end{tabular}}
\caption{The results over Homologous datasets.}
\label{table:in-domain}
\end{table}

Compared with the current methods, our method outperforms in all settings. Through a comparative analysis of existing methods, we further illustrate the reasons why our approach shows significant advantages (Table \ref{table:in-domain}).

Adversarial training like DANN may fail in some cases, which coincides with the results by \cite{wright-augenstein-2020-transformer} in the domain adaptation. Although DANN adversarially trains a domain classifier to make features indistinguishable, the domain confounders may well also remain and the feature suffers from the spurious correlation. Our method of invariant representation via backdoor adjustment well tackles this issue.

Data augmentation method like Mixup fails in most cases, as the gain in random augmentation does not cover distribution difference triggered by domain shift. Augmented data more closely to the target domain \cite{yu-etal-2021-cross, calderon-etal-2022-docogen} may be more effective than the universal method, but this is not possible when the target domain is not accessible.

Other methods focusing on learning invariant prediction (like GroupDRO, VREx, EQRM) fall behind ours because they don't consider domain-specific information. It also demonstrates the sufficiency of our consideration of domain-specific information.

\subsection{Robustness}
\label{sect:robustness}
\begin{figure}
    \centering
    \includegraphics[scale=0.23]{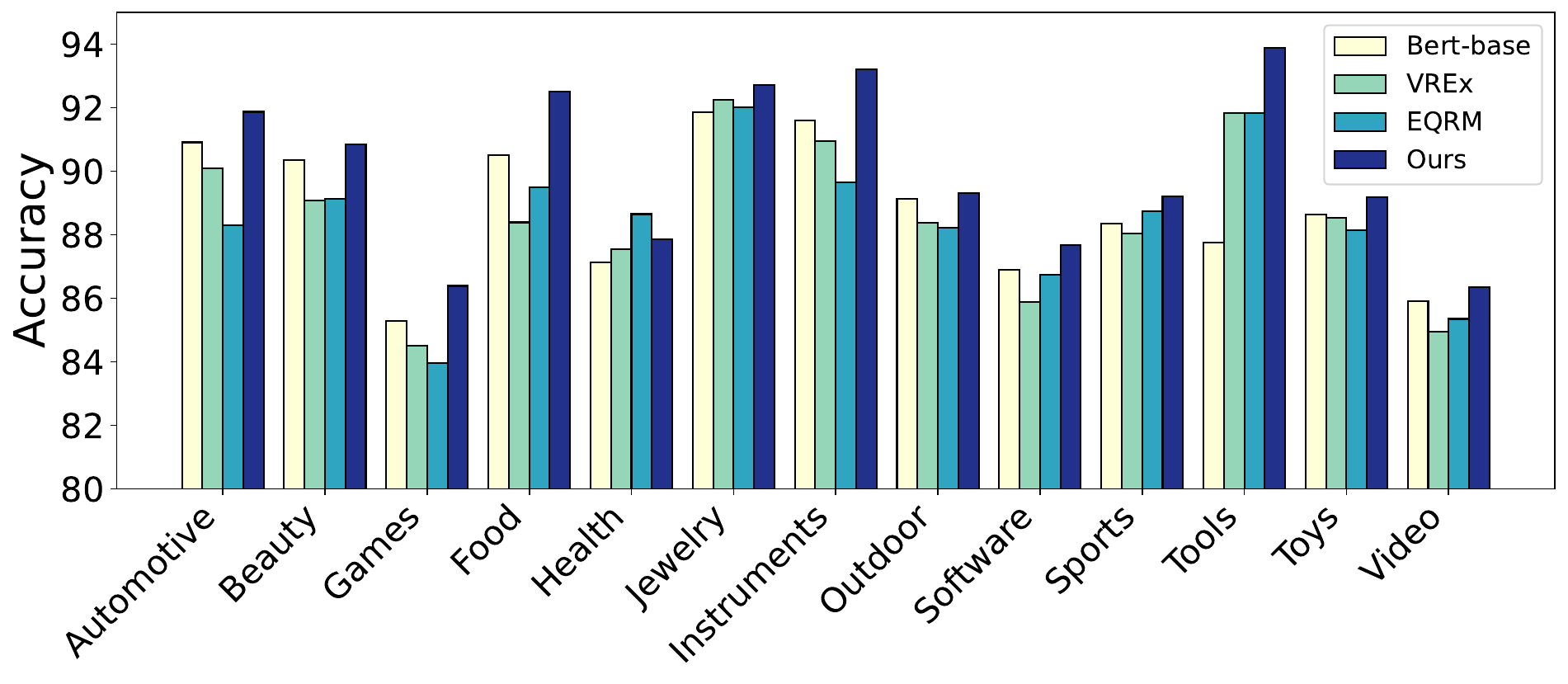}
    \caption{Results on 13 Homologous datasets.}
    \label{fig:robustness}
\end{figure}

To further evaluate the robustness of our model, we conduct experimental results on more homologous datasets of Amazon (Figure \ref{fig:robustness}). 
Specifically, we train our model on four domains (i.e., Books, DVDs, Electronics, and Kitchen appliances) and test on 13 other unseen domains. 
It is evident from the results that our method improves the performance of Bert-base in generalizing to multiple unseen domains. 
Specifically, our model outperforms the Bert-base model over all 13 domains.
These denote that our model has a wider generalization capability as it has the causal ability to reason both domain-invariant and domain-specific features.

\subsection{Performance on Diverse Domain} 
\label{sect:Diversity Domain}
Considering the major research in cross-domain sentiment analysis base on the amazon benchmark and only transfer across the products. We consider a more challenging setting on Diverse Dataset, where the distribution gap between the source domain and target domain is much bigger, like the restaurant domain to the airline domain rather than DVDs to Electronics. We maintain the same training setting of homologous datasets, three domain data for training and validation, and one domain for testing (Table \ref{table:out-domains}). 
For example, Airline in Table \ref{table:out-domains} means training on Amazon, IMDb, and Yelp, and testing on Airline.

In contrast to the homologous scenario, the performance of different test domains shows relatively large differences, reflecting the difference between the fields as well.
Due to growing differences between domains and smaller common features, the gain from the original method decreases (like VREx). Owing to our capability to causally model both invariant and specific information, our approach is still able to maintain good performance in the diverse scenario.

\begin{table}[t!]
\centering
\small
\setlength{\tabcolsep}{0.8mm}{\begin{tabular}{lccccc}
\hline
          & Airline       & Amazon         & Imdb           & Yelp          & Avg   \\ \hline
Bert-base & 82.02          & 87.70           & 88.05          & 94.85          & 88.16 \\
DANN      & 82.62          & 87.90          & 88.45               &   93.80          & 88.19  \\
Mixup     & 82.67         & 87.85         &  89.65          & 94.65         & 88.71  \\
GroupDRO  &     83.17	&     88.65       &  87.95     & 94.15          & 88.48  \\
IRM  &     83.32	&     88.60       &  89.35     & 92.75          & 88.51  \\
VREx      &  83.37	   & 88.75       &  88.05       &  94.20              & 88.59      \\ 
EQRM      &  83.47	   & 87.75       &  89.20       &  94.35              & 88.69      \\ 
\hline
Ours      & \textbf{85.81} & \textbf{89.10} & \textbf{90.00} & \textbf{95.20} & \textbf{90.03}  \\ 
Improvement &  +2.34    & +1.35      &  +0.80      & +0.85  & +1.34 \\ \hline
\end{tabular}}
\caption{The performance over diverse datasets.}
\label{table:out-domains}
\end{table}

\subsection{Further Analysis}
\label{sect:further analysis}
\paragraph{Ablation Studies}
To further analyze the effectiveness of the key parts in our model, we provide the ablation studies (Table \ref{table:ablation study}). Specifically, we remove the backdoor loss and adjustment loss that aims to learn a domain-invariant representation (w/o Invariant), the specific loss and joint loss that is designed to learn a domain-specific representation (w/o Specific), and both of them (w/o Both (Bert-base)). 
From the results, we can obtain the following findings.
First, the backdoor adjustment can help our model learn a better domain-invariant representation, which improves the generalization of our model (row 1 and 2). 
Second, both domain-invariant and domain-specific representations are important for cross-domain sentiment analysis (row 1-3). Removing one of them will reduce the performance of all four datasets.

\paragraph{Representation Visualization}
To better understand our model, we visualize the representation of the text (Figure \ref{fig:representation visualization}) over two homologous and diverse datasets. 
Particularly, we compare our model with Bert-base model by obtaining the representations of samples in the test set.
we use t-SNE to translate the 768-dimension representation into a 2-dimension vector. 
We can observe that it is hard for Bert-base to distinguish the samples with different sentiment polarities. 
In contrast, the gap between our invariant representations of positive and negative samples is clear. 
These observations indicate that our backdoor adjustment helps our model learn good generalized representations.

\begin{table}[t!]
\centering
\small
\setlength{\tabcolsep}{0.8mm}
{\begin{tabular}{lcccc}
\hline
                                    & DEK-B          & EKB-D          & KBD-E          & BDE-K          \\ 
\hline
ours                                & \textbf{90.20} & \textbf{90.15} & \textbf{90.95} & \textbf{91.95} \\
w/o Invariant & 88.30          & 90.00          & 88.30          & 89.85          \\
w/o Specific         & 89.75          & 89.65          & 90.05          & 89.15          \\
w/o Both (Bert-base)                           & 88.10          & 89.65          & 90.00          & 90.35          \\ 
\hline
\end{tabular}}
\caption{The ablation results of our model. }
\label{table:ablation study}
\end{table}

\newcommand{\mysize}{3.6 cm}
\begin{figure*}[!htb]
    \centering
    \subfigure[electronics(bert-base)]{\includegraphics[width=\mysize]{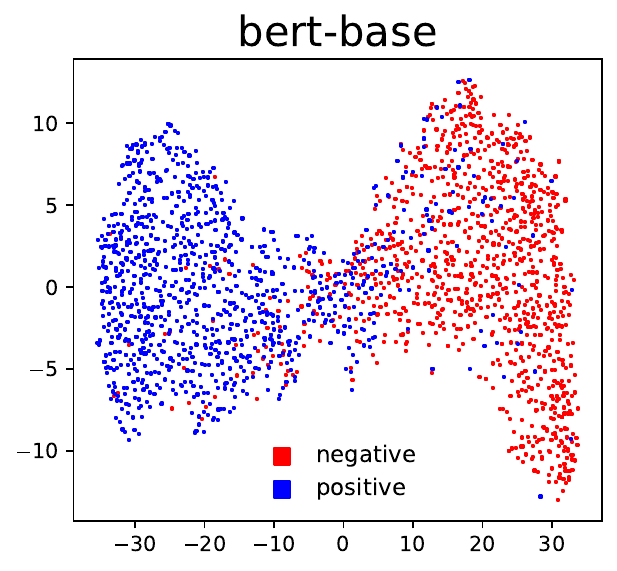}\label{fig: sub_figure1_1}}
    \subfigure[electronics(ours)]{\includegraphics[width=\mysize]{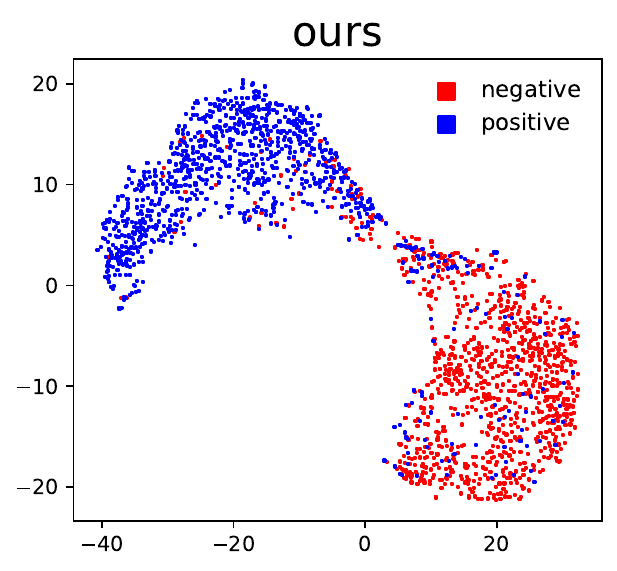}\label{fig: sub_figure1_2}}
     \subfigure[kitchen(bert-base)]{\includegraphics[width=\mysize]{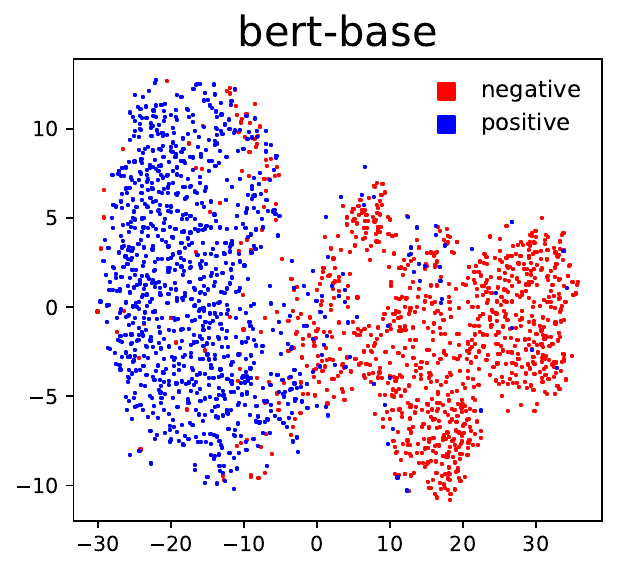}\label{fig: sub_figure2_1}}\
    \subfigure[kitchen(ours)]{\includegraphics[width=\mysize]{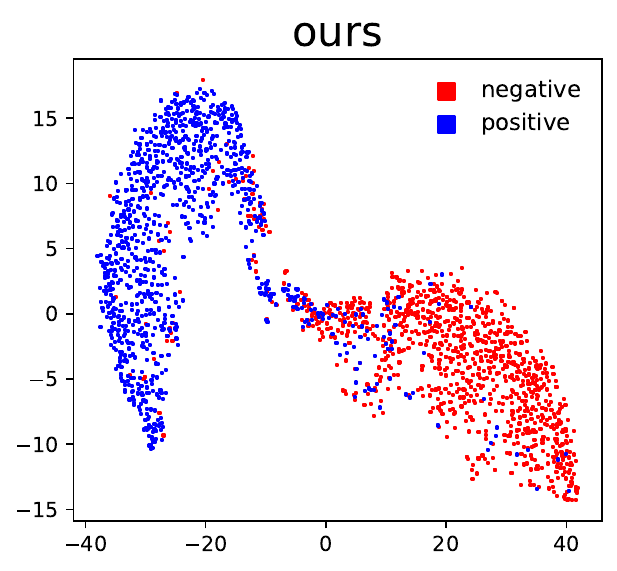}\label{fig: sub_figure2_2}}
    
    \subfigure[yelp(bert-base)]{\includegraphics[width=\mysize]{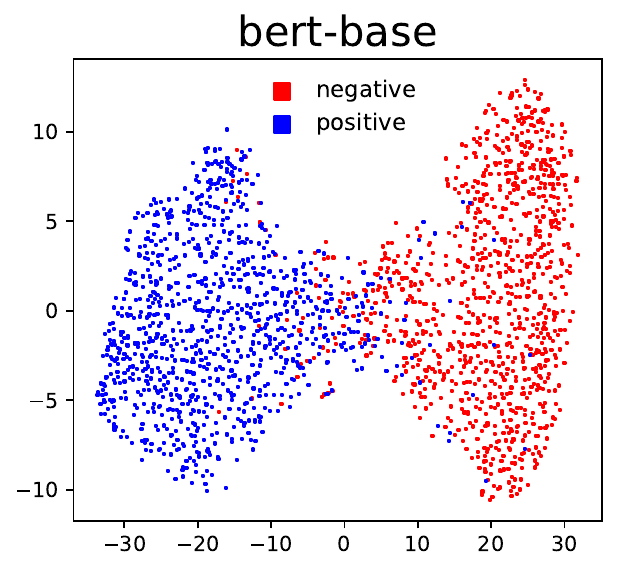}\label{fig: sub_figure3_1}}
    \subfigure[yelp(ours)]{\includegraphics[width=\mysize]{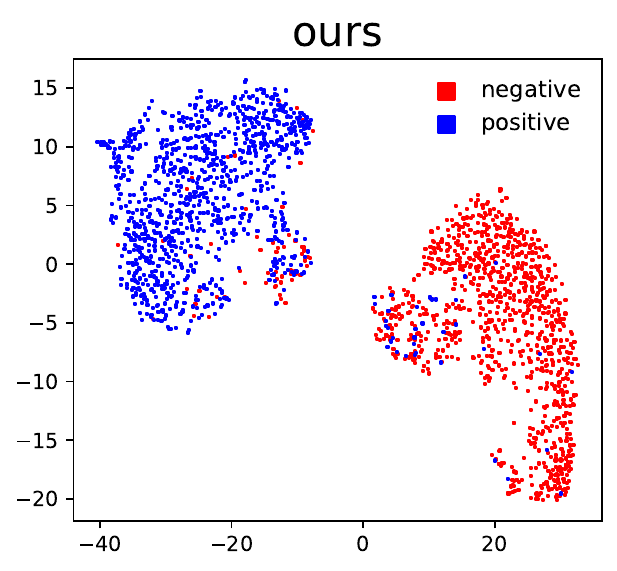}\label{fig: sub_figure3_2}}
    \subfigure[amazon(bert-base)]{\includegraphics[width=\mysize]{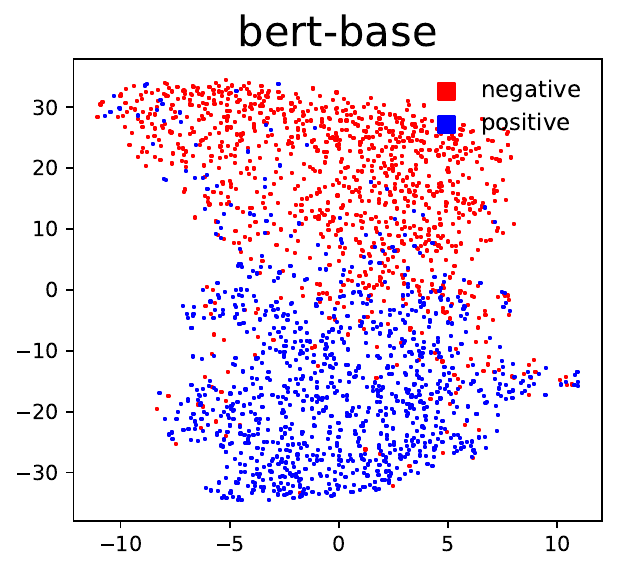}\label{fig: sub_figur4_1}}
    \subfigure[amazon(ours)]{\includegraphics[width=\mysize]{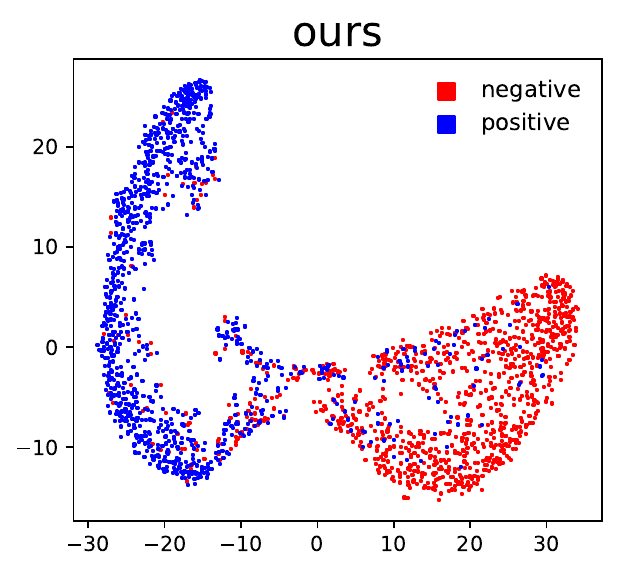}\label{fig: sub_figure4_2}}
    \caption{Representation visualization.}
    \label{fig:representation visualization}
\end{figure*}

\paragraph{Comparison with Large Language Models (LLMs)}

To comprehensively assess the implications of our research, we have undertaken a performance evaluation comparing our model with ChatGPT, a well-established Large Language Model (LLM) (Table \ref{tab:llm_comparison}). The experiments involved evaluating both models using zero-shot and few-shot techniques across three diverse datasets: books, musical instruments, and tools \& hardware. 
We gauged the effectiveness of our model in comparison to ChatGPT under various scenarios, shedding light on the relative strengths and weaknesses of our approach against this established LLM.

While ChatGPT exhibits superior performance in the book domain, attributable to its pre-training on extensive datasets encompassing common domains (e.g., books), our model demonstrates competitive efficacy. Notably, our model outperforms ChatGPT in categories such as musical instruments and tools \& hardware. This suggests that, although LLMs excel in specific domains due to their pre-training on large-scale datasets, they may encounter challenges in generalizing beyond those domains. The observed differences highlight the intricate nature of language models, underscoring the critical importance of addressing and resolving the fundamental challenge of domain generalization, even for large-scale models.

\begin{table}[t!]
\centering
\setlength{\tabcolsep}{1.0mm}
\begin{tabular}{lccc}
\hline
  & B & M & T \\
\hline
ERM (Bert-base) & 88.10 & 91.59 & 87.76 \\
ChatGPT (zero-shot) & 91.89 & 89.64 & 86.73 \\
ChatGPT (3-shot) & \textbf{93.86} & 92.88 & 91.84 \\
Ours & 90.20 & \textbf{93.20} & \textbf{93.88} \\
\hline
\end{tabular}
\caption{Performance Comparison with LLM (ChatGPT)- (B) Book, (M) Musical Instruments, (T) Tools \& Hardware}
\label{tab:llm_comparison}
\end{table}



\section{Related Work}
\label{sec:related}

\subsection{Cross-Domain Sentiment Analysis}
Cross-domain sentiment analysis aims to generalize a classifier that is trained on a source domain, for which typically plenty of labeled data is available, to a target domain, for which labeled data is scarce \cite{blitzer-etal-2007-biographies, du2020adversarial}. 
In \textbf{single-source domain adaptation}, one line of work employs pivot features to bridge the gap between a source domain and a target domain \cite{blitzer-etal-2007-biographies, Yu2016LearningSE, Ziser2018PivotBL, peng2018cross, BenDavid2020PERLPD}.
Another branch expects to learn invariant representation, by adversarial training \cite{ganin2016domain, Li2017EndtoEndAM, du2020adversarial}, contrastive learning \cite{Long2022DomainCC}.
Motivated by the success of masked pre-trained language models, some other recent studies base on data augmentation \cite{calderon-etal-2022-docogen, Wang2022CounterfactualRA}, prompt tuning \cite{Wu2022AdversarialSP}.

There is also a little research on \textbf{multi-source domain adaptation}, which uses multiple domains as sources and adapts to one target domain. In this setting, limited works mainly focus on adversarial training \cite{Zhao2018AdversarialMS, Chen2018MultinomialAN, Wu2019DualAC} and mixture of expert \cite{guo-etal-2018-multi}. 
However, most domain adaptation models assume access to unlabeled data from the target domain in-hand during training. For a more reasonable generalization setting, we consider a more challenging and realistic setting, \textbf{domain generalization},  where only source domain data can be used during training.

\subsection{Domain Generalization}
In recent years, domain generalization (DG) has received much attention in ML, which can be divided into three categories \cite{wang2022generalizing}: 
(1) \textbf{Data Manipulation. } 
Data manipulation/augmentation methods \cite{DBLP:conf/iclr/ZhangCDL18, DBLP:conf/coling/SunXYLYH20} aim to increase the diversity of existing training data with operations including randomization, transformation, etc.
(2) \textbf{Invariant representation learning. }
A widely used method is adversarial training \cite{DBLP:conf/icml/GaninL15, Li2018DomainGW}, which adversarially trains the generator (to fool the discriminator) and discriminator (to distinguish the domains). 
In other works, learning invariant features is approximated by enforcing some invariance conditions across training domains by adding a regularization term to the usual empirical risk minimization \cite{arjovsky2019invariant, DBLP:conf/icml/KruegerCJ0BZPC21}. 
Some group-based works \cite{sagawa2019distributionally, Liu2021JustTT} improve the worst group performance.
(3) \textbf{Learning Strategy. }
This line of work focuses on exploiting the general learning strategy to promote the generalization capability, like meta-learning \cite{chen2022discriminative}, and ensemble learning \cite{Mancini2018BestSF, guo-etal-2018-multi}. 


There are also several works that consider extending DG to the NLP field, including rumor detection and MNLI \cite{BenDavid2021PADAAP}, SLU \cite{Shen2021EnhancingTG}, Text-to-SQL\cite{Gan2021ExploringUL}, 
Semantic Parsing \cite{Marzinotto2019RobustSP, Wang2020MetaLearningFD}, etc. In this work, we consider the DG for cross-domain sentiment analysis. 
Following the line of invariant representation learning, we propose the backdoor condition for invariant representation and balance the domain-specific features.

\subsection{Causality for NLP}
Recent years have witnessed the boom of causality, many research combines causal inference with existing machine learning approaches to achieve good results \cite{Feder2021CausalII}. This method is used in a wide range of fields, including spurious correlation \cite{wang-culotta-2020-identifying, Veitch2021CounterfactualIT, wang-etal-2022-causal}, data augmentation \cite{Zmigrod2019CounterfactualDA, Liu2021CounterfactualDA}, interpretability \cite{Vig2020CausalMA, Elazar2020AmnesicPB}, etc.
The most related to our method is several works utilizing backdoor adjustment to debias in various NLP application, including text classification \cite{Landeiro2016RobustTC}, distantly supervised named entity recognition \cite{zhang-etal-2021-de}, court's view generation \cite{Wu2020DebiasedCV}. 
Different from the current method using backdoor adjustment only in the inference period , we design the backdoor adjustment as an invariant prediction condition and add it into the training period to achieve the invariant representation.

\section{Conclusion}
\label{sec:conclusion}
In this paper, we consider a more challenging scenario, domain generalization for cross-domain sentiment analysis, where the target domain is unseen.
Therefore, we propose a framework that disentangles domain-invariant and domain-specific features and leverages both to predict.
We rethink the cross-domain sentiment analysis in a causal view and uncover the potential confounders in so-called invariant representations.
Taking inspiration from the backdoor adjustment in causal intervention, we propose the backdoor condition to achieve an invariant representation that is not confounded by the domain.
Extensive experimental results on more than 20 homologous and diverse datasets demonstrate the great generalization of our model in cross-domain sentiment analysis.

\section*{Limitations}
Experimental results show that our proposed invariant representation learning does alleviate the problem of potential confounders triggered by domain.
Despite giving some examples of domain knowledge as the confounder, it remains impossible for us to enumerate in detail all confounder cases and how variations in performance improvement vary with the differences in confounder between domains.
We expect more good interpretable approaches to unveil the potential confounders in cross-domain sentiment analysis and explain the validity of our proposed invariant learning satisfying the backdoor condition. 

Moreover, as with other DG studies, the hyperparameters need to be set manually, which limits generalization to some extent.
In the future, we expect to eliminate this manual process through self-learning, etc., so that the model is more generalizable.

\section*{Acknowledge}
The authors wish to thank the reviewers for their helpful comments and suggestions. This work was partially funded by National Natural Science Foundation of China (No.62376061).

\nocite{*}
\section*{Bibliographical References}\label{sec:reference}

\bibliographystyle{lrec-coling2024-natbib}
\bibliography{lrec-coling2024-example}

\section*{Language Resource References}
\label{lr:ref}
\bibliographystylelanguageresource{lrec-coling2024-natbib}
\bibliographylanguageresource{languageresource}

\end{document}